\pdfoutput=1

\documentclass[11pt]{article}

\usepackage{acl}

\usepackage{times}
\usepackage{inconsolata}
\usepackage{latexsym}

\usepackage[T1]{fontenc}

\usepackage[utf8]{inputenc}

\usepackage{listings}
\usepackage{microtype}
\usepackage{graphicx}
\usepackage{amsmath}
\usepackage{bbm}
\usepackage{amssymb}
\usepackage{amsfonts}
\usepackage{booktabs}
\usepackage{lipsum}
\usepackage{multicol}
\usepackage{multirow}
\usepackage{misra}
\usepackage{tikz}
\usepackage{xcolor}
\usepackage{caption}
\usepackage{subcaption}
\usepackage{enumitem}
\usepackage{graphicx}
\usepackage[most]{tcolorbox}

\newcommand{\minicons}{\texttt{minicons}}
\newcommand{\transformers}{\texttt{transformers}}
\newcommand{\scorer}{\textbf{\texttt{scorer}}}
\newcommand{\cwe}{\textbf{\texttt{cwe}}}
\newcommand{\lmzoo}{\texttt{lm-zoo}}
\newcommand{\jiant}{\texttt{jiant}}
\newcommand{\diagnose}{\texttt{diagNNose}}
\newcommand{\blimp}{\textsc{BLiMP}}
\newcommand{\multiberts}{Multi\textsc{Bert}s}

\newcommand{\anli}{$\alpha$NLI}
\DeclareMathOperator*{\argmax}{argmax}

\definecolor{Code}{rgb}{0,0,0} 
\definecolor{Decorators}{rgb}{0.5,0.5,0.5} 
\definecolor{Numbers}{rgb}{0.5,0,0} 
\definecolor{MatchingBrackets}{rgb}{0.25,0.5,0.5} 
\definecolor{Keywords}{rgb}{0,0,1} 
\definecolor{self}{rgb}{0,0,0} 
\definecolor{Strings}{rgb}{0,0.63,0} 
\definecolor{Comments}{rgb}{0.63,0,0} 
\definecolor{Backquotes}{rgb}{0,0,0} 
\definecolor{Classname}{rgb}{0,0,0} 
\definecolor{FunctionName}{rgb}{0,0,0} 
\definecolor{Operators}{rgb}{0,0,0} 
\definecolor{Background}{rgb}{0.99,0.99,0.99} 

\definecolor{cadmiumgreen}{rgb}{0.0, 0.42, 0.24}
\definecolor{oldmauve}{rgb}{0.4, 0.19, 0.28}
\definecolor{royalazure}{rgb}{0.0, 0.22, 0.66}
\definecolor{harvardcrimson}{rgb}{0.79, 0.0, 0.09}
\definecolor{lightmauve}{rgb}{0.86, 0.82, 1.0}
\definecolor{darkbrown}{rgb}{0.4, 0.26, 0.13}

\usepackage{setspace}
\title{\texttt{minicons}: Enabling Flexible Behavioral and Representational Analyses of Transformer Language Models}

\author{Kanishka Misra\\
  Purdue University \\
  \texttt{kmisra@purdue.edu}}

\begin{document}
\maketitle
\begin{abstract}
We present \texttt{minicons}, an open source library that provides a standard API for researchers interested in conducting behavioral and representational analyses of transformer-based language models (LMs). 
Specifically, minicons enables researchers to apply analysis methods at two levels: (1) at the prediction level---by providing functions to efficiently extract word/sentence level probabilities; and (2) at the representational level---by also facilitating efficient extraction of word/phrase level vectors from one or more layers.
In this paper, we describe the library and apply it to two motivating case studies: One focusing on the learning dynamics of the BERT architecture on relative grammatical judgments, and the other on benchmarking 23 different LMs on zero-shot abductive reasoning. minicons is available at \url{https://github.com/kanishkamisra/minicons}.
\end{abstract}

\section{Introduction}
Accessing and using pre-trained language models (LMs)---now a mainstay in modern NLP research---has become ever so convenient due to the advent of high-quality open source libraries such as \transformers{} \citep{wolf-etal-2020-transformers}, \jiant{} \citep{pruksachatkun-etal-2020-jiant}, etc.
Parallel to the proliferation of newer pre-training methods and LM architectures is that of the development of analyses methods and diagnostic datasets which we collectively refer to here as the field of `BlackboxNLP' \citep{alishahi2019analyzing}. One of the foundational goals of this field is to develop and understanding of exactly what is learned by these complex LM architectures as a result of pre-training, as well as how pre-trained models operate.

\begin{figure}[t]
    \centering
    \includegraphics[width=\columnwidth]{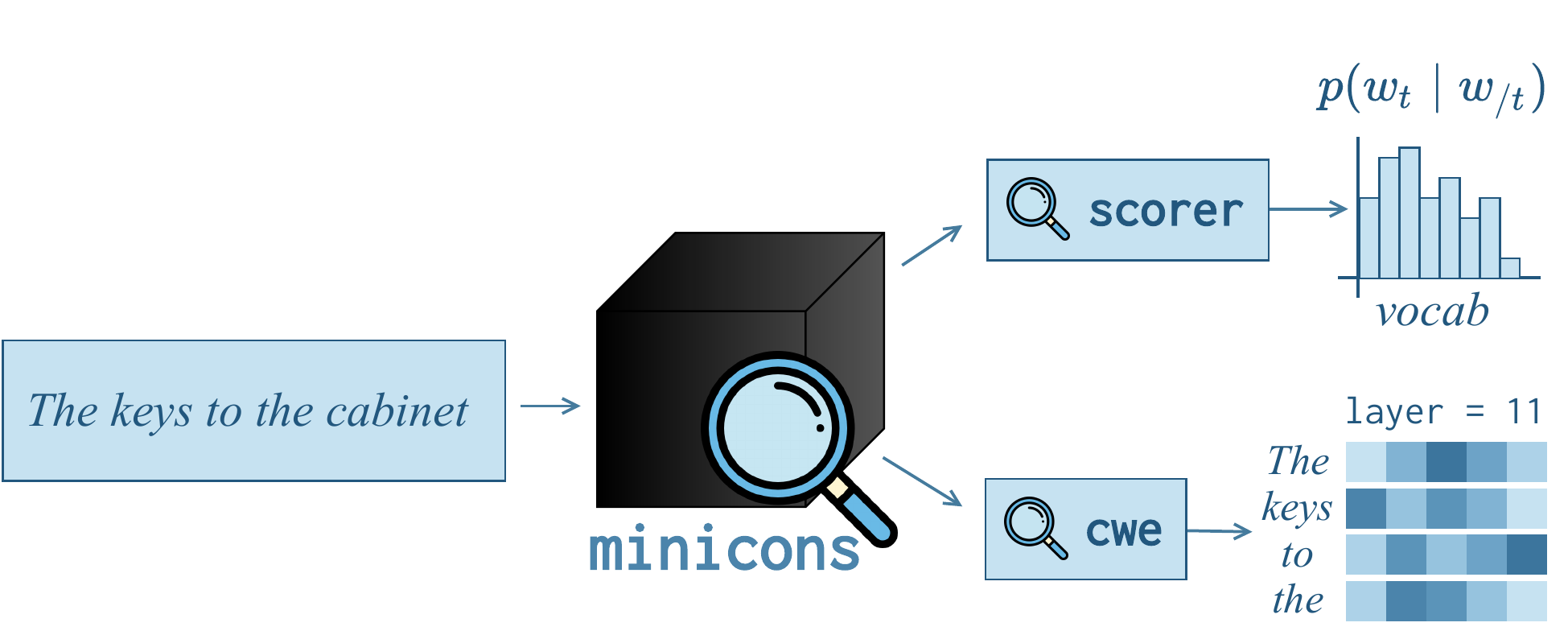}
    \caption{\minicons{} facilitates analyses of pre-trained transformer-based language models at the \textbf{prediction} level through its \scorer{} module and at the \textbf{representational} level through its \cwe{} module. Detailed overview provided in \cref{sec:overview}.}
    \label{fig:my_label}
\end{figure}
In this paper, we provide an implementation-level solution to conduct such analyses in the form of an open-source python library called \minicons{}. \minicons{} builds on top of the \transformers{} library \citep{wolf-etal-2020-transformers}, and provides a standard API to perform behavioral and representational analyses of transformer \citep{vaswani2017attention} language models. In particular, it contains methods that collectively support analyses at two levels, both of which are well established in current LM analysis and evaluation literature: (1) at the prediction level, by using LM's word prediction abilities to perform a range of analyses that target specific linguistic or reasoning ability without performing any supervised training/fine-tuning; and (2) at the representational level, where the goal is to characterize the information made available in the internal activations of a given model. Together, these methods make standard model analyses methods accessible to a wider audience.

In what follows, we first provide an overview of minicons and its core modules. We then apply it in two motivating case studies. First, we analyze the learning dynamics of the BERT architecture with respect to 67 different grammatical phenomena related to syntax, semantics, and morphology. Second, we use \minicons{} to measure the extent to which the patterns in LMs' sequence probabilities align with human-like abductive commonsense reasoning, the task of providing the most plausible explanation given partial observations -- shedding light on their capacity to make abductive inferences in a `zero-shot' manner.

\section{Overview of \texttt{minicons}\footnote{Documentation found at: \url{https://minicons.kanishka.website}}}
\label{sec:overview}
\subsection{Dependencies and Requirements}
\minicons{} is released under the MIT License on \texttt{pypi}, and can be installed using the bash command: \texttt{pip install minicons}. It supports python 3.7 or later, and is built on top of \texttt{pytorch} \citep{paszke2019pytorch} and \texttt{transformers} \citep{wolf-etal-2020-transformers}. Therefore, it can be used for \textit{any} LM that is accessible through the HuggingFace model hub. \minicons{} can run on both CPUs and GPUs. All computations in the two modules can be performed batch-wise, making \minicons{} suitable for large-scale behavioral analyses.

\subsection{Core Modules}
\minicons{} has two core modules, each of which facilitates two types of model analyses, as discussed below. In addition, we provide example code for using these modules in \Cref{app:code}.
\paragraph{\texttt{scorer}}
A number of behavioral analyses of LMs focus on investigating them in their natural environment---the task of estimating word probabilities in context.
Such analyses typically elicit scores that correspond to word/sequence probabilities or their mathematical modifications (such as surprisals) and use them for different investigations including (but not limited to): Evaluation of their capacity to judge relative linguistic acceptability \citep[][etc.]{linzen2016assessing, marvin-linzen-2018-targeted, futrell-etal-2019-neural, warstadt-etal-2020-blimp-benchmark}, even in a cross/multi-lingual setting \citep{mueller-etal-2020-cross}; Statistical relationship between LM word probabilities and human behavioral/neurological data such as self-paced reading times \citep{merkx-frank-2021-human}, EEG/MEG readings \citep{hollenstein-etal-2020-zuco}, etc.; Assessment of models' commonsense, semantic, and pragmatic knowledge \citep[][etc.]{ettinger-2020-bert, shwartz-etal-2020-unsupervised}, as well as ability to pick up societal biases \citep{nangia-etal-2020-crows} in unsupervised settings.
\minicons{} supports all such analyses through the \textbf{\texttt{scorer}} module. 
This module includes utilities for two predominant estimators of word probabilities in context -- (1) Masked language models (instantiated by the \textbf{\texttt{scorer.MaskedLMScorer}} class), and standard autoregressive language models (instantiated by the \textbf{\texttt{scorer.IncrementalLMScorer}} class). It defines equivalent functions to elicit different kinds of word prediction data from both these model classes -- \textbf{\texttt{token\_score()}} for word level scores, \textbf{\texttt{sequence\_score()}} for sentence-level scores, and a hybrid \textbf{\texttt{partial\_score()}} for sequence scores in cases where either of the conditioned or the predicted items are held constant \citep[as used by][among others]{nangia-etal-2020-crows, misra-etal-2020-exploring, holtzman-etal-2021-surface, misra2021typicality}. 
By default, these methods produce log-probabilities, but can also additionally elicit modified scores such as surprisals \citep[a key predictor of reading times, see][]{smith2013effect} as well as the rank of each word in context \citep [important for investigations based on common sense or world knowledge, for example,][]{petroni-etal-2019-language}. 
Finally, the \scorer{} module can also be accessed using a command line interface (see \Cref{sec:cli}).

\paragraph{\texttt{cwe}} While the scorer module aids in behavioral analyses at the prediction level, the \cwe{} module allows researchers to shed light on the information made available at the representation level. The primary function of this module is to provide a standard API to extract ``contextualized embeddings''---activations of the neural network from various layers of a transformer LM using the \textbf{\texttt{cwe.extract\_representation()}} function. One can either extract embeddings from one layer at a time or even combine layers as needed \citep[as used by][for instance]{loureiro-jorge-2019-language}, and can also extract phrase or sentence level embeddings (in which case the module also accepts a reduction method, which by default is an average over sub-word embeddings). Like the \scorer{} module, this module also allows batched computation, which can be further sped up by using GPUs. Broadly, the \cwe{} module facilitates representational analysis methods such as (but not limited to): probing for various linguistic competencies using \textit{probing classifiers} \citep[e.g.,][among others]{ettinger2016probing, conneau2018you, hewitt-manning-2019-structural}; extraction and evaluation of sense-embeddings using sense-annotated corpora \citep[e.g.,][]{loureiro-jorge-2019-language, nair-etal-2020-contextualized} to facilitate lexical-semantic analyses; Representational Similarity Analysis \citep{kriegeskorte2008representational} between models and across model layers \citep{abnar-etal-2019-blackbox}.

\subsection{Relation to other Libraries}
\minicons{} is closely related to two other libraries that focus on model analysis and evaluation -- \lmzoo{} \citep{gauthier-etal-2020-syntaxgym}, which provides a command line interface to access word prediction statistics from 7 pre-trained LMs; and \diagnose{} \citep{jumelet-2020-diagnnose}, which provides methods to perform activation extraction, feature attribution, and targeted syntax evaluations of pre-trained LSTMs and transformer LMs.
\minicons{} extends the coverage of LM-based scoring methods provided by both these libraries by also incorporating the MLM-scoring algorithm, and therefore allows access to a broader set of pre-trained LMs. Furthermore, unlike \diagnose{}, \minicons{} is not restricted to fixed (although commonly used) templates of model analyses and instead places the control of output complexity in the hands of the user -- for instance, one can choose between three modes of LM-elicited probability measurements: token, sequence, or partially conditioned. Similarly, \minicons{} also allows for custom logic to reduce sub-word/multi-word representations---e.g. taking the mean, or selecting the head-word, etc---a feature missing from both the aforementioned libraries.

\section{Motivating Case Studies\footnote{code for reproducing these experiments can be found at \url{https://github.com/kanishkamisra/minicons-experiments}}}
To demonstrate the usefulness of \minicons{}, we focus on two motivating case studies that feature heavy use of the library's \scorer{} module.
We focus only on this module here, as the \cwe{} module is primarily a utility tool for extracting contextual representations of words, phrases, and sentences, and therefore only supplies inputs to more sophisticated representational analysis methods.\footnote{nevertheless, readers interested in demonstrations using the \cwe{} module can refer to \href{https://minicons.kanishka.website/representations.html}{\texttt{this url}} or \Cref{app:code}.}
By contrast, the \scorer{} module allows end-to-end analyses and is self-contained.

\subsection{Learning Dynamics of Relative Linguistic Acceptability in LMs}
\label{sec:multiberts}
Our first case study involves using \minicons{} to test LMs' knowledge of linguistic acceptability, the task of judging whether a given sentence is acceptable under the rules of a given language \citep{lawrence2000natural, lau2017grammaticality, warstadt-etal-2019-neural}.
To do so, we follow the common paradigm of providing LMs with minimal pairs---sentences that usually differ in one or two words---of acceptable and unacceptable sentences, and evaluating the extent to which the model prefers the correct sentence as acceptable \citep{warstadt-etal-2020-blimp-benchmark}.
For instance, when provided the minimal pairs in example \cref{ex:blimpex1}, a linguistically competent model with knowledge of number agreement should prefer \cref{ex:good1} over \cref{ex:bad1}.

\ex.\label{ex:blimpex1} \textsc{Anaphor Agreement (number)}
    \a.\label{ex:good1} These patients do respect \underline{themselves}.
    \b.\label{ex:bad1} *These patients do respect \underline{himself}.

Specifically, in this section, we use the tools provided by \minicons{} to shed light on how knowledge required to assess relative linguistic acceptability emerges and evolves during the course of an LM's training.
We do so by evaluating the LM at various time steps as it is pre-trained on its word-prediction-based objective.
Such an inquiry can supplement contemporary analyses of linguistic acceptability LMs, which usually focus on model performance \textit{after} pre-training, and paint a more comprehensive picture of model behavior as it learns to predict words in context.
\paragraph{Data} We use as our source of minimal pairs the \blimp{} benchmark \citep{warstadt-etal-2020-blimp-benchmark}, perhaps the largest and most fine-grained dataset of its kind.
\blimp{} covers 67 different linguistic paradigms/tasks, each of which has 1000 minimal pairs like example \cref{ex:blimpex1}. The various linguistic paradigms are further classified into 12 different linguistic phenomena \citep[see table 4 in][]{warstadt-etal-2020-blimp-benchmark}, each belonging to either \textit{syntax}, \textit{semantics}, both \textit{syntax} and \textit{semantics}, or \textit{morphology}.
\paragraph{Models} We evaluate the learning dynamics of linguistic acceptability in \multiberts{} \citep{sellam2021multiberts} -- reproduced variants of the \texttt{bert-base-uncased} model \citep{devlin-etal-2019-bert}, trained on the same corpora (wikipedia and BookCorpus) for 2M steps, using different seeds.
We specifically evaluate the 28 checkpoints\footnote{during training, a checkpoint is saved every $20{,}000$ steps up to the $200{,}000^{\text{th}}$ step, and thereafter every $100{,}000$ steps.} released by the authors for models trained using five different seeds, amounting to 145 different \texttt{bert-base-uncased} models ($5 \times 28$ checkpoints $+$ $5$ initial, untrained models). 
We compare our results to the original \texttt{bert-base-uncased} model \citep{devlin-etal-2019-bert}.

\paragraph{Method}
The dominant paradigm of LM evaluation using minimal pairs is to subject the LM with a forced-choice task: an LM correctly selects the acceptable sentence in a minimal pair if it assigns a higher likelihood to it.
Since our model checkpoints and the reference model are all bidirectional masked LMs, we use the \textbf{\texttt{scorer.MaskedLMScorer}} class to instantiate them. We then use its \textbf{\texttt{sequence\_score()}} method to compute \textit{pseudo-loglikelihoods} as the approximation of the log-probabilities of the input batch of sentences, following the MLM-scoring method proposed by \citet{salazar-etal-2020-masked}.
In our computations, we further divide the pseudo-loglikelihoods by the number of tokens in the input to control for the difference in sentence lengths \citep{lau2017grammaticality}.
The accuracy of a model for a given phenomenon is then simply the percentage of times it correctly assigns the acceptable sentence ($A$) higher probability relative to the unacceptable sentence ($U$).
That is, for a dataset $\mathcal{D} = \{(A_1, U_1), \dots, (A_n, U_n)\}$ containing stimuli for a given phenomenon, we calculate the model's accuracy as:
\begin{align}
    \frac{1}{n}\sum_{i=1}^{n}\mathbbm{1}\underbrace{\left\{\frac{\log p_\theta(A_i)}{\mid A_i\mid} > \frac{\log p_\theta(U_i)}{\mid U_i \mid}\right\}}_{\text{computed using \textbf{\texttt{sequence\_score()}}}},
\end{align}
where $\mathbbm{1}$ is the indicator function, which returns 1 if its condition is met; otherwise, 0.
We calculate this measure per linguistic phenomenon in \blimp{}, for every \multiberts{} checkpoint, as well as the original BERT-base model.

\begin{figure*}[!t]
    \centering
    \includegraphics[width=\textwidth]{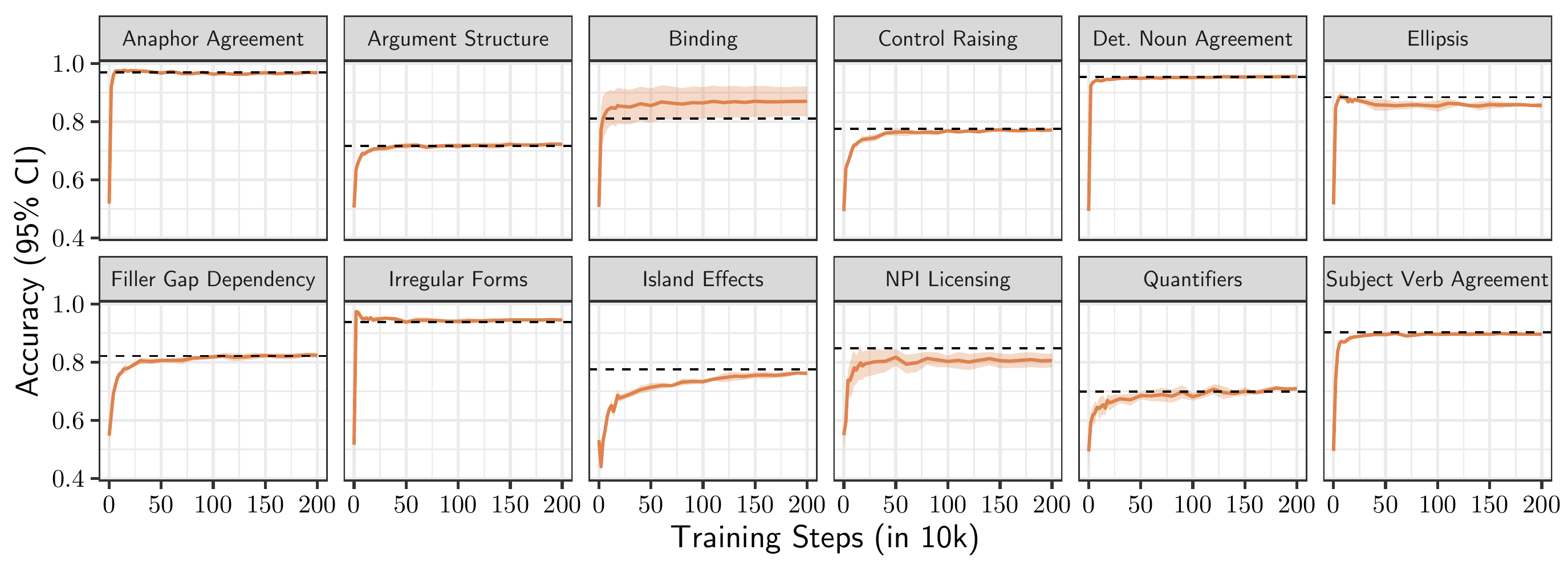}
    \caption{Learning dynamics of \multiberts{} \citep{sellam2021multiberts} on various \blimp{} phenomena plotted as accuracy over training time-steps. Error-bands represent 95\% confidence intervals, while the dashed line represents performance of \texttt{bert-base-uncased} \citep{devlin-etal-2019-bert}. Chance performance is 0.50 throughout.}
    \label{fig:multiberts}
\end{figure*}

\paragraph{Analysis and Results}
The results of our \blimp{} experiments are shown in \Cref{fig:multiberts}. 
In this figure, we plot the accuracy of the various \multiberts{} for each of the 12 \blimp{} phenomena at various stages of their training and compare them to the accuracy of the original BERT-base model \citep{devlin-etal-2019-bert}, which is fully trained and therefore shows constant performance for each phenomenon.
For most phenomena, we find the learning dynamics of \multiberts{} to eventually converge to the performance of the original BERT model. However, the rate at which they do so differs slightly based on the linguistic phenomena in question.
In particular, phenomena based on number and gender agreement are learned reasonably early during training, and converge with BERT-base as early as 20{,}000 training steps.
These are followed by Argument Structure, Ellipsis, and Irregular forms, which are then followed by the rest (see \cref{fig:multiberts}). Island effects are learned the slowest, suggesting that this capacity is acquired gradually compared to other phenomena. 
Interestingly, the performance of \multiberts{} on phenomena involving Ellipsis and Irregular forms degrades slightly after reaching the level of BERT-base early on, indicating mild signs of ``forgetting'' during MLM training -- where presumably the features responsible for capturing knowledge of Irregular morphology and Ellipsis are slightly degraded after peaking in the first 20{,}000 and 60{,}000 steps, respectively.
The performance of \multiberts{} in NPI Licensing is consistently below that of BERT-base, while that on Binding phenomena is almost always better than BERT-base, showing consistent improvements as early as 40{,}000 steps and remaining constant thereafter, reaching its peak at 1.5M steps with an accuracy of 0.87, 6 percentage points above BERT-base performance. We also observe a great amount of variability in these phenomena as opposed to the others, suggesting that BERT's capacity to encode these phenomena is highly sensitive to the initial weights of the model. Finally, all models relatively struggle on paradigms involving knowledge of Quantifier and Argument structure.

Our results complement those of \citet{zhang-etal-2021-need}, who focus on learning curves of the RoBERTa-base, which are computed based on the amount of training data---quantified as the number of tokens observed during training. Although \citet{sellam2021multiberts} do not report token-level statistics of \multiberts{}, a similar analysis can be readily performed using \minicons{}.

\subsection{Unsupervised Abductive Natural Language Inference}
The capacity of LMs to estimate sequence probabilities lends itself well to zero-shot and unsupervised analyses and benchmarks that focus on ``commonsense reasoning'' \citep{shwartz-etal-2020-unsupervised, bigbench, klein-nabi-2021-towards}. For example, instances of the Winograd Schema Challenge can be reinterpreted in a zero-shot setting by supplying an LM a prompt such as \textit{``The trophy could not fit in the suitcase because it was too big. What was too big?''} and comparing the relative probabilities (conditioned on the prompt) of \textit{trophy} and \textit{suitcase} as elicited by the LM to perform evaluation.
This line of work is increasingly gaining traction, as it sheds light on the extent to which statistical reasoning based on complicated co-occurrence statistics---an ability presumably encoded as a result of pre-training---can result in predictions that are consistent with the ones made by employing more sophisticated and human-like reasoning processes.

Motivated by the rise in evaluations concerning zero-shot/unsupervised ``reasoning'' using LM-based sequence probabilities, we use \minicons{} to analyze several pre-trained LMs on their ability to perform \textit{abductive} reasoning -- the capacity to make inferences to the most plausible explanation, given a set of observations \citep{peirce1974collected}. Inferences made using abductive reasoning are necessarily probabilistic and do not focus on deductive truth (unlike in standard entailment tasks).
In our analyses we compare the zero-shot performance of various LMs on the \textit{abductive natural language inference} task \citep[\anli;][]{bhagavatula2019abductive}.
An instance of the \anli{} task provides
observations that occur at different times: $O_1$ at time $t_1$, and $O_2$ at time $t_2 > t_1$. It further includes two hypotheses $H_1$ and $H_2$ that serve as candidate explanations for the two observations. The task, then, is to select the hypothesis that is more plausible given the two observations. An example is given below, with the most plausible hypothesis emboldened:

\begin{tcolorbox}[colback=blue!5,colframe=blue!75!black]
\begin{enumerate}[align=left,nolistsep]
    \item[$O_1$:] Tim was entering a baking contest.
        \item[$H_1$:] \textit{Tim made an extremely greasy donut.}
        \item[$H_2$:] \textbf{\textit{Tim made a great cheese cake.}}
    \item[$O_2$:] Tim won the baking contest.
\end{enumerate}
\end{tcolorbox}

\begin{figure*}[t]
    \centering
    \begin{subfigure}[b]{0.70\textwidth}
        \centering
        \includegraphics[width=\textwidth]{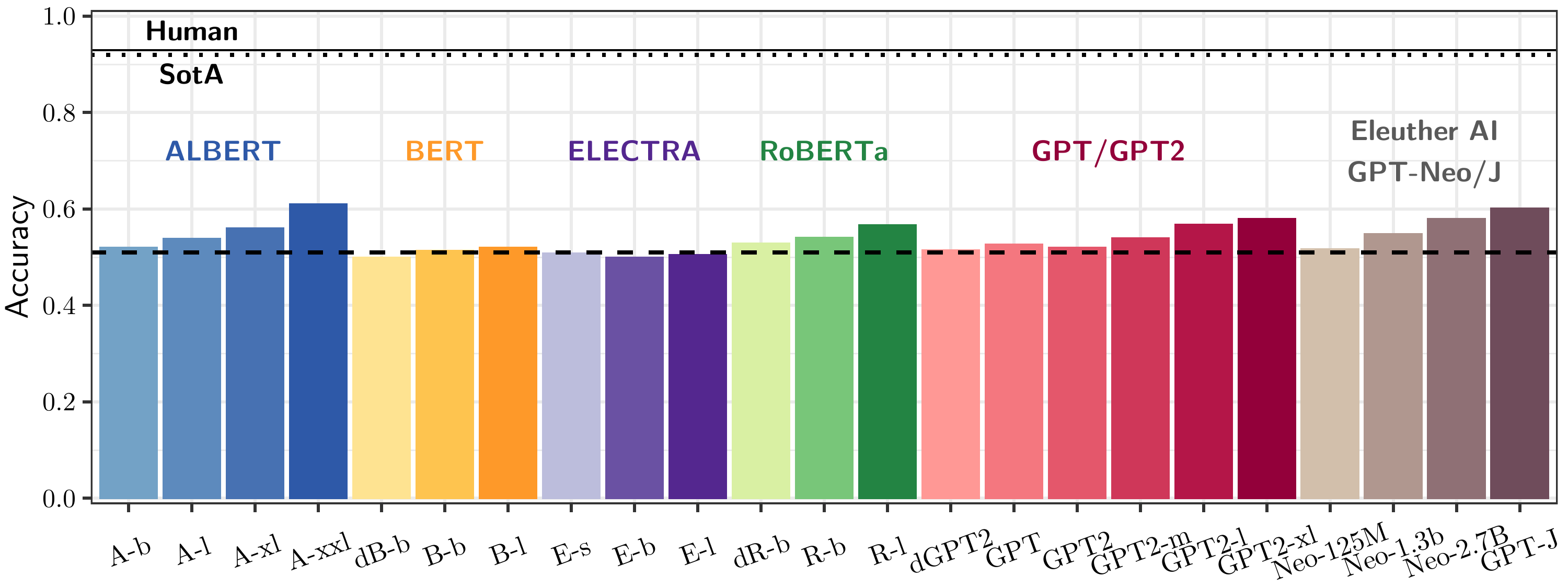}
        \caption{}
        \label{fig:anliperformance}
    \end{subfigure}
    \hfill
    \begin{subfigure}[b]{0.29\textwidth}
        \centering
        \includegraphics[width=\textwidth]{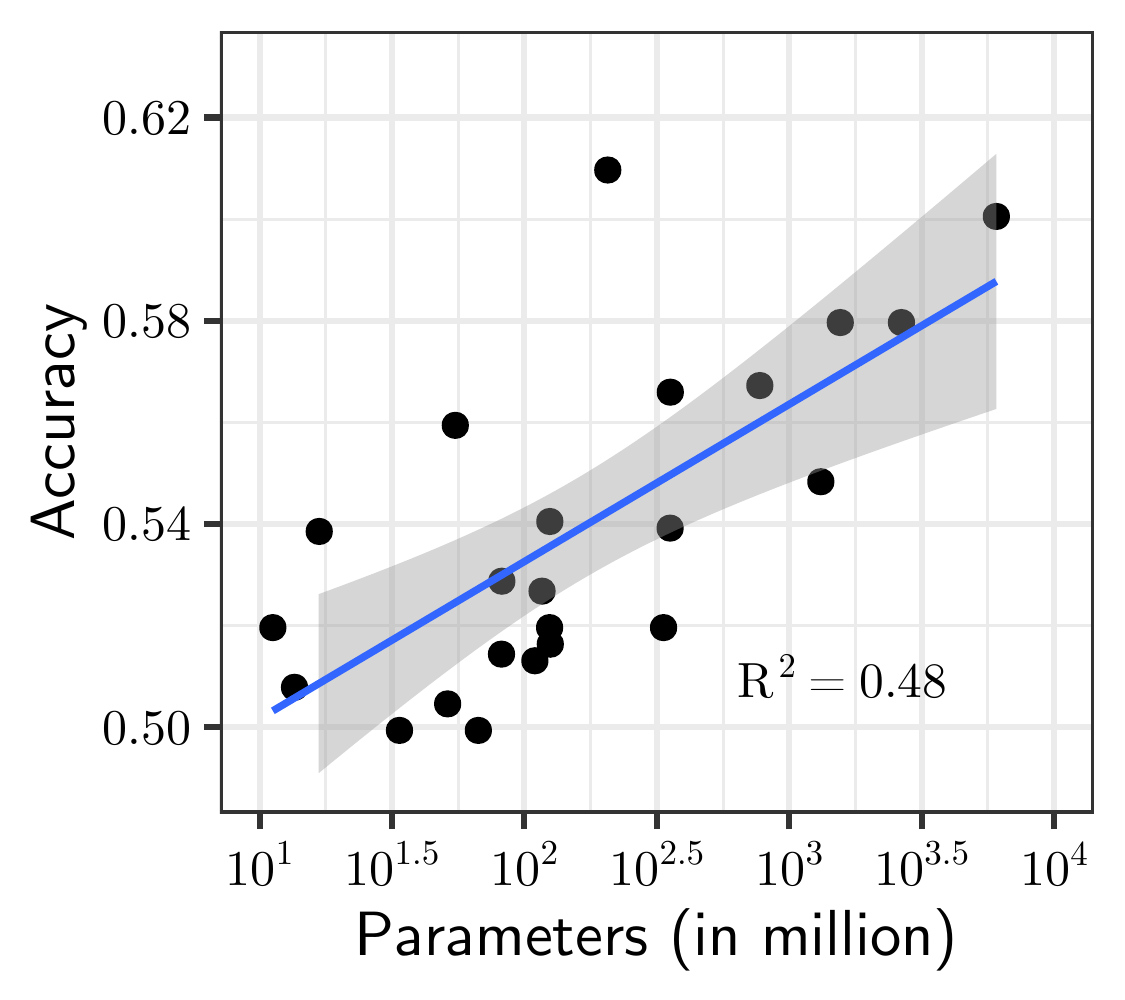}
        \caption{}
        \label{fig:anlicorrelation}
    \end{subfigure}
    \caption{Performance of the Unsupervised LM-scoring method on the \anli{} development set \citep{bhagavatula2019abductive}. \textbf{(a)} Accuracy of each of the 23 models, arranged and colored based on the model family. Chance performance is 0.51 (dashed line), while current state of the art and human performance are 0.92 and 0.93, respectively (shown in dotted and solid lines, respectively). \textbf{(b)} Scatter plot of model accuracies versus parameter count (in millions; log-scaled). $R^2 = 0.48, p < .01$.}
    \label{fig:anli}
\end{figure*}

\paragraph{Data} We use the \anli{} dataset \citep{bhagavatula2019abductive}, and evaluate on the development set.
\paragraph{Models} 
We benchmark six different LM families--four masked LM architectures: (1) BERT \citep{devlin-etal-2019-bert}, (2) RoBERTa \citep{liu2019roberta}, (3) ALBERT \citep{lan2019albert}, (4) ELECTRA \citep{clark2020electra}; and two autoregressive LMs: (1) GPT \citep{radford2018improving} and GPT2 \citep{radford2019language}; (2) GPT-Neo \citep{gpt-neo} and GPT-J \citep{gpt-j}, together considered under the `EleutherAI' family. Additionally, we used distilled versions of the BERT, RoBERTa, and GPT2 architectures, trained using the method described in \citet{sanh2019distilbert}. This results in a total of 23 different pre-trained LMs, all of which were accessed from the Hugging Face Hub.\footnote{\url{https://huggingface.co/models}} A comprehensive summary of these models, including total parameters, tokenization scheme, and corpus sizes, is shown in \Cref{tab:modelmeta} (see \Cref{sec:modelmeta}).

\paragraph{Method} 
Following recent work in unsupervised commonsense reasoning using pre-trained LMs \citep{shwartz-etal-2020-unsupervised, holtzman-etal-2021-surface}, we use sequence log-probabilities to benchmark the abductive reasoning capacities in our 23 LMs. More specifically, given an instance of the \anli{} dataset, $(O_1, H_1, H_2, O_2)$, we select the hypothesis that maximizes the conditional probability $p(O_2 \mid O_1, H)$. That is,
\begin{align}
    H_+ = \argmax_{i \in \{1,2\}}\underbrace{\log p_\theta(O_2 \mid O_1, H_i)}_{\text{computed using \textbf{\texttt{partial\_score()}}}},
\end{align}
where $H_+$ is the predicted hypothesis.
This operates under the assumption that the hypothesis that best explains the given observations sequentially follows $O_1$ and precedes $O_2$ -- i.e., a hypothesis is the more plausible explanation (out of the two) given $O_1$ if it more strongly leads an LM to generate $O_2$. A similar assumption is made by \citet[][section 3]{bhagavatula2019abductive}.
We compute our conditional probabilities using the \textbf{\texttt{partial\_score()}} function for our LMs, which are instantiated using either the \textbf{\texttt{scorer.MaskedLMScorer}} or the \textbf{\texttt{scorer.IncrementalLMScorer}} class, depending on their architecture. This function computes the conditional log-probability of sequences based on either pseudo or standard language model scoring.
Using this method, we evaluate our 23 models on the development set provided by \citep{bhagavatula2019abductive} based on their accuracy, and submit the predictions of the best performing model on the \anli{} leaderboard\footnote{\url{https://leaderboard.allenai.org/anli/submissions/public}} to get the test set performance.

\paragraph{Analysis and Results}
\Cref{fig:anli} shows the results of applying the above method to the 23 pretrained LMs. Overall, we find most models to be at or slightly above chance performance and far below human-level and state of the art performance, obtained using fine-tuning and data-augmentation techniques (see \Cref{fig:anliperformance}). 
Interestingly, we find ALBERT-xxlarge-v2 \citep{lan2019albert} achieves the best performance of the 23 models (accuracy = 0.61), despite being $\approx$30 times smaller than the largest model (GPT-J, accuracy = 0.60) in terms of total parameters. 
This highlights its surprising parameter efficiency, which is notable considering that a majority of Masked LMs are close to chance-level performance (e.g., all models in the BERT and ELECTRA families). 
This model achieved a test set accuracy of 0.61 on the \anli{} leaderboard, outperforming the best unsupervised model (named `GPT2-medium-unsupervised') by 3.5 percentage points, and being only 2.2 percentage points short of the performance obtained by fine-tuning BERT-base. 
From \Cref{fig:anlicorrelation}, we find that the performance on the development scales linearly with the logarithm of the number of parameters ($R^2 = 0.48$) of the models. This suggests that drastic improvements in unsupervised LM-based abductive reasoning are unlikely to arise from model scaling,\footnote{a naive extrapolation from our analyses suggests an increase of model expressivity by 5 orders of magnitude to reach close to state of the art performance.} but rather from more nuanced transformations -- a promising line of work in this regard is to incorporate explicit commonsense knowledge into the reasoning process \citep[e.g. like in][]{shwartz-etal-2020-unsupervised}. In general, our results highlight the difficulty of performing LM-based abductive reasoning in a zero-shot setting.

\section{Conclusion}
This paper presented \minicons{}, a utility tool to facilitate analyses of transformer-based language models based on their of-the-shelf behavior on controlled stimuli as well as on the information that their representations encode as a result of their training on large corpora. Through its integration with the ever-growing Hugging Face Model hub, \minicons{} is also suitable to run large-scale benchmarking experiments.
\minicons{} is an evolving project and we hope to integrate newer utility functions into the library as well as develop detailed tutorials to explain various analysis pipelines to new users. We especially welcome and encourage open source contributions to the library.

\paragraph{Acknowledgments} \minicons{} has benefited tremendously from valuable discussions with Hemanth Devarapalli, Forrest Davis, and Sanghee J. Kim, as well as from its active users. The author thanks Bruno Nicenboim and Adele Goldberg, whose use of the package in its initial stages revealed embarrassingly obvious bugs. The experiments reported in this paper were partially run on the Gilbreth cluster at Purdue University's Rosen Center for Advanced Computing, and partially run on Hemanth Devarapalli's computational platform with an NVIDIA 3090 GPU. Finally, thanks to Julia Taylor Rayz for allowing the use of \minicons{} in her NLP class at Purdue University.

\bibliography{custom}
\bibliographystyle{acl_natbib}
\clearpage

\appendix

\section{Command Line Interface}
\label{sec:cli}
\begin{figure}[h]
    \centering
    \includegraphics[width=\columnwidth]{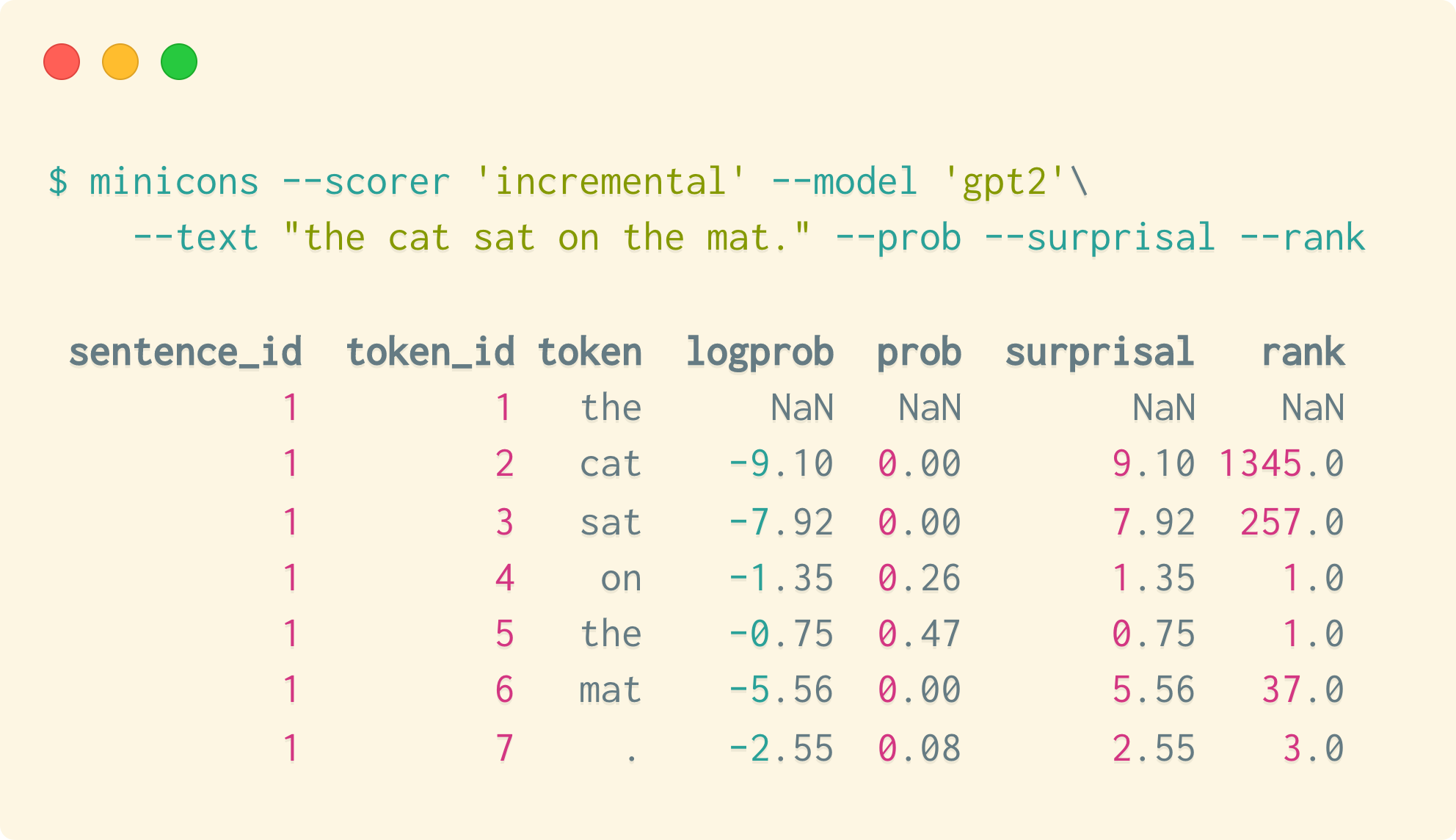}
    \caption{An example usage of the \minicons{} CLI, showing results when GPT2 \citep{radford2019language} is used to elicit word-level scores from the sentence \textit{the cat sat on the mat.} Scores for first word are \texttt{NaN} as the first token is used to initiate the conditioning for the rest of the sentence.}
    \label{fig:cli}
\end{figure}

The \minicons{} library is accompanied by a command-line interface (CLI) that can be used to readily elicit word or sentence level scoring using any pre-trained transformer LM that is accessible on the huggingface hub or saved on the user's local directory. \Cref{fig:cli} shows an example usage of the CLI.

\section{Code Samples}
\label{app:code}
We illustrate the use of \minicons{} using the following snippets (not exhaustive):
\begin{itemize}
    \item Listing 1 illustrates the use of the \scorer{} module to compute word and sequence level probabilities using GPT2. The stimuli being compared are similar to those used in the \blimp{} experiments.
    \item Listing 2 illustrates the use of the \cwe{} module to extract contextual word representations at one or more layers of \texttt{bert-base-uncased}.
    \item Listing 3 illustrates the use of the \scorer{} module to compute and query the output distributions of masked LMs (in principle, can also be done for Autoregressive models, although only for the last token). The \textbf{\texttt{query\_vocab}} function is especially useful in analyses such as that of \citet{newman-etal-2021-refining}, where one wants to compare the probabilities of words present in a predefined list.
\end{itemize}

\renewcommand{\figurename}{Listing}

\begin{lstlisting}[float=*,caption=Using the \scorer{} module to compare grammatical and ungrammatical sentences. The variables \texttt{good} and \texttt{bad} can in principle be batches of sentences. This is essentially a pseudo-code for the \blimp{} experiment in \cref{sec:multiberts}.]
from minicons import scorer

# Instantiate model (here, gpt2)
lm = scorer.IncrementalLMScorer('gpt2')

# define grammatical and ungrammatical sentences
good ='The key to the cabinet is on the table.'
bad = 'The keys to the cabinet is on the table.'

# compute by-token log-probabilities
good_score = lm.sequence_score(good)
bad_score = lm.sequence_score(bad)

good_score
#> [-4.218889236450195]

bad_score
#> [-4.223154067993164]

# get token-level scores:
lm.token_score(good)
#>[[('The', 0.0),
#>  ('key', -7.177484512329102),
#>  ('to', -1.3995437622070312),
#>  ('the', -2.6050567626953125),
#>  ('cabinet', -10.562873840332031),
#> ... the rest.
\end{lstlisting} 

\begin{lstlisting}[float=*,caption=Using the \cwe{} module to extract contextualized word representations from different layers of an LM.]
from minicons import cwe

# Instantiate model (here, bert-base)
lm = cwe.CWE('bert-base-uncased')

# First way of representing stimuli:
# [sentence, word]
stimuli = [
    ['The robin flew away.', 'robin'], 
    ['Robin is my favorite bird.', 'Robin']
]

# Alternate way of representing input stimuli:
# [sentence, character_span]
stimuli = stimuli = [
    ['The robin flew away.', (4, 9)], 
    ['Robin is my favorite bird.', (0, 5)]
]

reps = lm.extract_representation(stimuli, layer = 11)
reps
#> tensor([[ 1.1954,  0.0493, -0.5261,  ..., -0.7852,  0.0137, -1.1233],
#>         [ 1.5843, -0.5463, -1.0030,  ..., -0.7533, -0.4128, -1.3711]])

reps.shape
#> torch.Size([2, 768])

# specifying multiple layers
reps = lm.extract_representation(stimuli, layer = [11, 12])
reps
#> [tensor([[ 1.1954,  0.0493, -0.5261,  ..., -0.7852,  0.0137, -1.1233],
#>          [ 1.5843, -0.5463, -1.0030,  ..., -0.7533, -0.4128, -1.3711]]),
#>  tensor([[ 1.0156,  0.0944, -0.8484,  ..., -0.3637,  0.1533, -0.6835],
#>          [ 0.8569, -0.5206, -1.0747,  ..., -0.3828,  0.5484, -0.3525]])]
\end{lstlisting} 

\begin{lstlisting}[float=*,caption=Using the \scorer{} module to compute and query the output distribution of \texttt{bert-base-uncased} for top-k predicted words\, as well as a forced-choice fill-in-the-blank task. The \textbf{\texttt{query\_vocab}} function also computes the rank of each forced-choice word.]
from minicons import scorer

# Instantiate model (here, bert-base-uncased)
lm = scorer.MaskedLMScorer('bert-base-uncased')

# specify masked token. [MASK] for bert-base
blank = lm.tokenizer.mask_token

# create stimuli using masked token.
stimuli = [
    f'Paris is the capital of {blank}.',
    f'Berlin is the capital of {blank}.'
]

stimuli = [[s, blank] for s in stimuli]

# get top-3 predictions:
lm.get_predictions(stimuli, k = 3)
#> [[('france', 0.950905442237854),
#>   ('algeria', 0.011535070836544037),
#>   ('morocco', 0.0032133283093571663)],
#>  [('germany', 0.6249764561653137),
#>   ('brandenburg', 0.24474896490573883),
#>   ('prussia', 0.05286181718111038)]]

# querying model with a restricted vocabulary (also shows rank based on prob.)
lm.query_vocab(stimuli, restricted_vocab = ['france', 'germany'])

#> [[('france', 0.950905442237854, 1), ('germany', 4.805942444363609e-05, 80)],
#>  [('france', 0.0005986356409266591, 22), ('germany', 0.6249764561653137, 1)]]
\end{lstlisting}

\section{Model Summaries}
\label{sec:modelmeta}
\Cref{tab:modelmeta} shows the model specifications for the 23 different LMs used in this paper (including \multiberts{}, which are essentially \texttt{bert-base-uncased} models trained with different seeds). All models were accessed using the huggingface hub.

\begin{table*}[h]
\centering
\resizebox{\linewidth}{!}{%
\begin{tabular}{clrrclr} 
\toprule
\textbf{Family} & \textbf{Model} & \textbf{Parameters} & \textbf{Vocab Size} & \textbf{Tokenization} & \textbf{Corpora} & \textbf{Tokens} \\ 
\midrule
\multirow{4}{*}{\textcolor[HTML]{2e59a8}{ALBERT}} & \texttt{albert-base-v2} & 11M & \multirow{4}{*}{30,000} & \multirow{4}{*}{SentencePiece} & \multirow{4}{*}{\textsc{Wiki} and \textsc{bc}} & \multirow{4}{*}{3.3B} \\
 & \texttt{albert-large-v2} & 17M &  &  &  &  \\
 & \texttt{albert-xl-v2} & 15M &  &  &  &  \\
 & \texttt{albert-xxl-v2} & 206M &  &  &  &  \\ 
\midrule
\multirow{3}{*}{\textcolor[HTML]{fe9929}{BERT}} & \texttt{distilbertbase-uncased} & 67M & \multirow{3}{*}{30,522} & \multirow{3}{*}{WordPiece} & \multirow{3}{*}{\textsc{Wiki} and \textsc{bc}} & \multirow{3}{*}{3.3B} \\
 & \texttt{bert-base-uncased} & 110M &  &  &  &  \\
 & \texttt{bert-large-uncased} & 345M &  &  &  &  \\ 
\midrule
\multirow{3}{*}{\textcolor[HTML]{54278f}{ELECTRA}} & \texttt{electra-small} & 13M & \multirow{3}{*}{30,522} & \multirow{3}{*}{WordPiece} & \multirow{2}{*}{\textsc{Wiki} and \textsc{bc}} & \multirow{2}{*}{3.3B} \\
 & \texttt{electra-base} & 34M &  &  &  &  \\
 & \texttt{electra-large} & 51M &  &  & \begin{tabular}[c]{@{}l@{}}\textsc{Wiki}, \textsc{bc}, \textsc{cw},\\\textsc{cc}, and \textsc{Giga}\end{tabular} & 33B \\ 
\midrule
\multirow{3}{*}{\textcolor[HTML]{238443}{RoBERTa}} & \texttt{distilroberta-base} & 82M & 50,265 & Byte-pair encoding & \textsc{owtc} & 2B \\
 & \texttt{roberta-base} & 124M & \multirow{2}{*}{50,265} & \multicolumn{1}{l}{\multirow{2}{*}{Byte-pair encoding}} & \multirow{2}{*}{\begin{tabular}[c]{@{}l@{}}\textsc{bc}, \textsc{cc-news},\\\textsc{owtc}, and \textsc{Stories}\end{tabular}} & \multirow{2}{*}{–} \\
 & \texttt{roberta-large} & 355M &  & \multicolumn{1}{l}{} &  &  \\ 
\midrule
\multirow{6}{*}{\textcolor[HTML]{93003a}{GPT/GPT2}} & \texttt{distilgpt2} & 82M & 50,257 & \multirow{6}{*}{Byte-pair encoding} & \textsc{owtc} & 2B \\
 & \texttt{gpt} & 117M & 40,478 &  & \textsc{bc} & 800M \\
 & \texttt{gpt2} & 124M & \multirow{4}{*}{50,257} &  & \multirow{4}{*}{\textsc{WebText}} & \multirow{4}{*}{–} \\
 & \texttt{gpt2-medium} & 355M &  &  &  &  \\
 & \texttt{gpt2-large} & 774M &  &  &  &  \\
 & \texttt{gpt2-xl} & 1.5B &  &  &  &  \\ 
\midrule
\multirow{4}{*}{\textcolor[HTML]{6f4c5b}{EleutherAI}} & \texttt{gpt-neo-125M} & 125M & \multirow{4}{*}{50,257} & \multirow{4}{*}{Byte-pair encoding} & \multirow{4}{*}{\textsc{pile}} & 300B \\
 & \texttt{gpt-neo-1.3B} & 1.3B &  &  &  & 380B \\
 & \texttt{gpt-neo-2.7B} & 2.7B &  &  &  & 420B \\
 & \texttt{gpt-j-6B} & 6B &  &  &  & 402B \\
\bottomrule
\end{tabular}
}
\caption{Summary of the models used in this paper. \multiberts{} \citep{sellam-etal-2020-learning} have the same specification as \texttt{bert-base-uncased}.\\ \textbf{Legend for Corpora:} \textsc{Wiki}: Wikipedia; \textsc{bc}: BookCorpus \citep{zhu2015aligning}; \textsc{cw}: ClueWeb \citep{callan2009clueweb09}; \textsc{cc}: CommonCrawl \textsc{Giga}: Gigaword \citep{graff2003english}; \textsc{owtc}: OpenWebTextCorpus \citep{Gokaslan2019OpenWeb}; \textsc{cc-news}: CommonCrawl News \citep{nagel2016ccnews}; \textsc{Stories}: Stories corpus \citep{trinh2018simple}; \textsc{WebText}: WebText corpus \citep{radford2019language}; \textsc{Pile}: The Pile \citep{gao2020pile}}
\label{tab:modelmeta}
\end{table*}

\end{document}